\pdfoutput=1
\documentclass[11pt]{article}

\usepackage[]{acl}

\usepackage{times}
\usepackage{latexsym}

\usepackage[T1]{fontenc}
\usepackage[utf8]{inputenc}

\usepackage{microtype}

\usepackage{inconsolata}

\usepackage{blindtext}
\usepackage{amsmath}
\usepackage{booktabs}
\usepackage{lscape} 
\usepackage{graphicx}
\usepackage{subcaption}
\usepackage{xspace}
\usepackage{arydshln}
\usepackage{bbm}
\usepackage[normalem]{ulem}
\usepackage{multirow}
\usepackage{array,tabularx}
\usepackage{todonotes}
\usepackage{amssymb}
\usepackage[acronym]{glossaries}
\usepackage{makecell}
\usepackage{textcomp}
\usepackage{rotating,amsmath,booktabs}
\usepackage{cleveref}
\usepackage{enumitem}
\usepackage{soul}

\newcommand{\llamatwoseven}{\textsc{Llama}2-7B\xspace}

\newcommand{\llamatwo}{\textsc{Llama}2-chat\xspace}
\newcommand{\llamathree}{\textsc{Llama}3-chat\xspace}
\newcommand{\mistral}{\textsc{Mistral}-7B-instruct\xspace}
\newcommand{\gpt}{\textsc{GPT}\xspace}
\newcommand{\gptthree}{\textsc{GPT}-3.5-turbo\xspace}
\newcommand{\gptfour}{\textsc{GPT}-4\xspace}

\newcommand{\llamatwosevenOnlyNr}{7B\xspace}

\newcommand{\llamatwoseventyOnlyNr}{70B\xspace}
\newcommand{\llamathreeeightOnlyNr}{7B\xspace}
\newcommand{\llamathreeseventyOnlyNr}{70B\xspace}

\newcommand{\cites}[1]{\citeauthor{#1}'s (\citeyear{#1})\xspace}

\newcolumntype{M}[1]{>{\arraybackslash}m{#1}}

\title{\textit{``I understand your perspective''}: LLM Persuasion through the Lens of Communicative Action Theory}

\author{Esra Dönmez  \and Agnieszka Falenska \\
    Institute for Natural Language Processing, University of Stuttgart\\
    Interchange Forum for Reflecting on Intelligent Systems, University of Stuttgart\\
    \normalsize{\texttt{esra.doenmez@ims.uni-stuttgart.de}}}

\begin{document}
\maketitle

\newacronym{nlp}{NLP}{Natural Language Processing}
\newacronym{lm}{LM}{Language Model}
\newacronym{lms}{LMs}{Language Models}
\newacronym{llms}{LLMs}{Large Language Models}
\newacronym{llm}{LLM}{Large Language Model}
\newacronym{ai}{AI}{Artificial Intelligence}
\newacronym{ml}{ML}{Machine Learning}
\newacronym{rlhf}{RLHF}{Reinforcement Learning from Human Feedback}
\newacronym{cai}{CAI}{Constitutional Artificial Intelligence}
\newacronym{gfh}{GfH}{Good for Humanity}
\newacronym{dl}{DL}{Deep Learning}
\newacronym{cmv}{CMV}{Change My View}

\begin{abstract}
Large Language Models (LLMs) can generate high-quality arguments, yet their ability to engage in \emph{nuanced and persuasive communicative actions} remains largely unexplored. This work explores the persuasive potential of LLMs through the framework of Jürgen Habermas' Theory of Communicative Action. It examines whether LLMs express illocutionary intent (i.e., pragmatic functions of language such as conveying knowledge, building trust, or signaling similarity) in ways that are comparable to human communication. 

We simulate online discussions between opinion holders and LLMs using conversations from the persuasive subreddit \emph{ChangeMyView}. We then compare the likelihood of illocutionary intents in human-written and LLM-generated counter-arguments, specifically those that successfully changed the original poster’s view. We find that all three LLMs effectively convey illocutionary intent---often more so than humans---potentially increasing their anthropomorphism. Further, LLMs craft sycophantic responses that closely align with the opinion holder's intent, a strategy strongly associated with opinion change. Finally, crowd-sourced workers find LLM-generated counter-arguments more \emph{agreeable} and consistently prefer them over human-written ones. These findings suggest that LLMs' persuasive power extends beyond merely generating high-quality arguments. On the contrary, training LLMs with human preferences effectively tunes them to mirror human communication patterns, particularly nuanced communicative actions, potentially increasing individuals' susceptibility to their influence.
\end{abstract}

\section{Introduction}
\label{introduction}
\begin{figure}[t]
    \centering
    \resizebox{.485\textwidth}{!}{
        \includegraphics{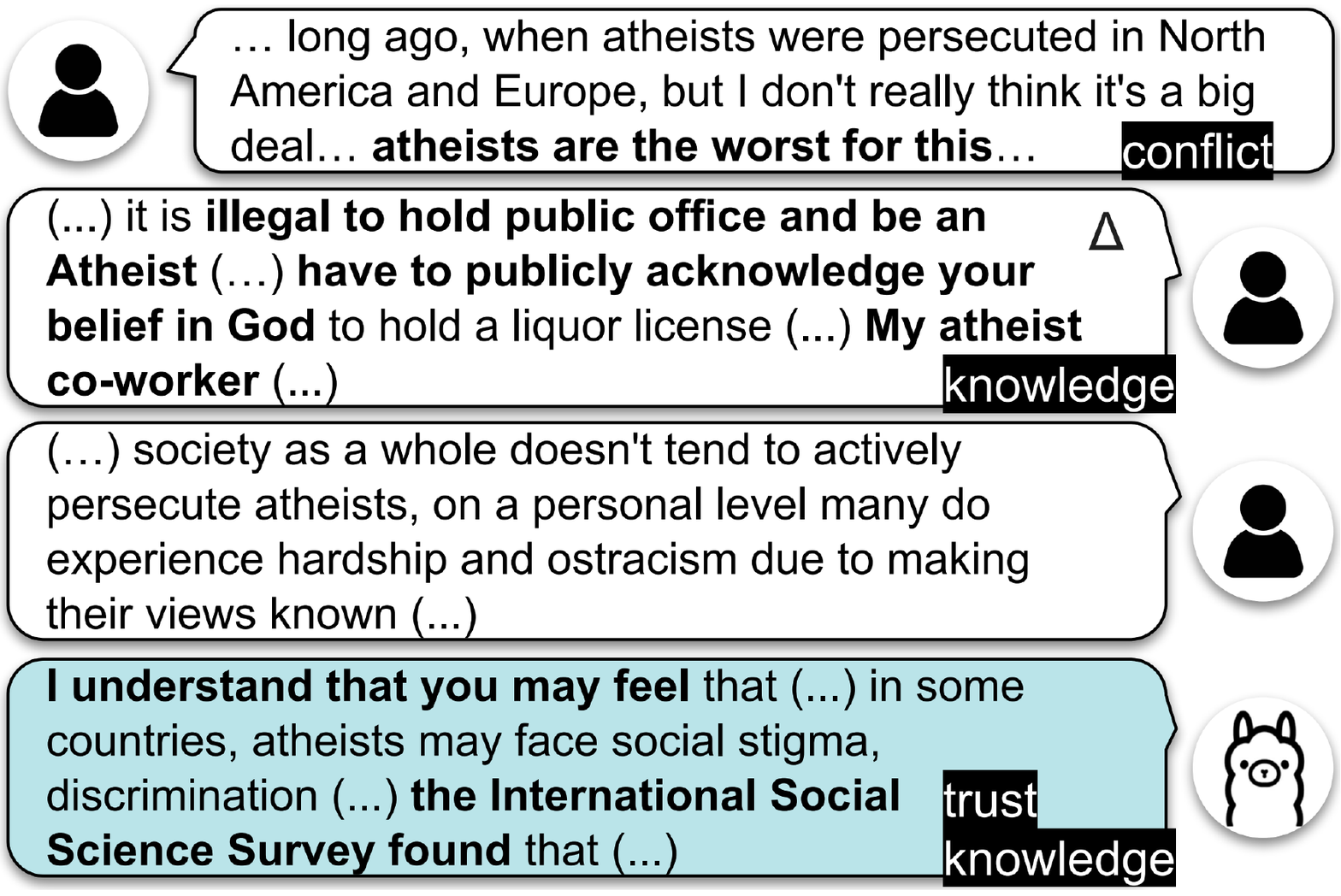}
    }
    \caption{Discussion from \emph{/r/ChangeMyView} subreddit annotated with pragmatic social dimensions (black boxes). Top: a post titled ``\emph{CMV: Atheists in Western nations aren't currently being persecuted or oppressed in any meaningful way}''. Below: two human-written comments (one opinion-changing, marked with $\Delta$). Bottom: a comment generated by Llama-2-7B.}
    \label{fig:example}
\end{figure}
Public discourse is essential for shaping opinions, exposing individuals to diverse perspectives, and challenging existing beliefs. Today, much of this discourse takes place online, especially on social media platforms \citep{monti2022language}, where AI-generated and edited content is becoming increasingly prevalent \cite{Hanley2024-machine-made-media}. \emph{What role does AI play in public discourse, particularly in shaping and changing opinions?}

Recent research highlights concerning trends in this regard, suggesting that humans may be susceptible to undue influence from LLMs on topics ranging from politics to social and environmental issues. A key challenge is that humans often struggle to recognize AI-generated text due to cognitive shortcuts, such as associating first-person pronouns or family-related topics with human authorship \cite{human_jakesch_2023}. Additionally, LLMs can infer the psychological dispositions of social media users \cite{large_peters_2024} and provide sycophantic responses (i.e., convincing, well-written responses that confirm a user’s mistaken beliefs), reinforcing confirmation bias \citep{sharma2024towards}.

Consider the conversation illustrated in \Cref{fig:example} (white boxes). A user expresses their opinion on atheists in Western nations using a \emph{conflicting} statement \textit{``atheists are the worst''}. Two users reply: the first shares \emph{knowledge} on the topic by listing facts and anecdotes, while the second simply states their opinion. Framing an argument within such illocutionary intents (\textbf{social dimensions}; see \citet{deri2018SocialDims} and further examples in \Cref{tab:social-dims}) can influence an interlocutor's opinion beyond the argument's quality alone \citep{monti2022language}. Effectively, language is not just a medium for conveying information but also a tool for exercising illocutionary force \cite{LAustin1962-LAUHTD} -- the capacity to influence a listener’s perspective and foster cooperation based on a shared understanding of reality. This is particularly effective when building ``\emph{shared knowledge, trust, and mutual understanding}'' \cite{Habermas1984-HABTTO}, which can drive opinion change and cooperation \cite{Krauss2000CommunicationAC, FRONZETTICOLLADON2023113317},  a process Habermas (\citeyear{Habermas1984-HABTTO}) terms Communicative Action.

\citet{monti2022language} examined the role of social dimensions in online persuasive discourse and found that counter-arguments containing at least one of the dimensions are more likely to change an individual's opinion. Additionally, matching the social dimension found in the original post -- such as responding to \emph{conflict} with \emph{conflict} -- increases the likelihood of persuasion. Clearly, social dimensions are strong indicators of persuasiveness yet remain underexplored in the context of LLMs. 
As exemplified in \Cref{fig:example} (bottom, blue box), models can also employ these strategies, potentially influencing readers' opinions in ways that remain unknown. Therefore, in this work\footnote{Our code \& data is available at
\url{https://github.com/esradonmez/llm_persuasion}}, we examine \textbf{the extent to which LLM-generated arguments contain social dimensions} and \textbf{how their expressed intent dynamics compare to human persuasive discourse on social media}. Specifically, we aim to address the following research questions:

\paragraph{RQ1} Which social dimensions are present in LLM-generated arguments, and how do they compare to human-written ones?

\paragraph{RQ2} What are the (dis-)similarities between the social dimensions expressed in successful human-written arguments and those generated by LLMs?

\paragraph{RQ3} To what extent can LLMs capture interactional dynamics of social dimensions between opinion holders and successful commenters?

\paragraph{RQ4} Do people prefer human-written opinion-changing arguments over LLM-generated ones? And do their preferences align more closely with the strategies used by humans or by LLMs?

To address these research questions, we simulate conversations between human opinion holders and LLMs using discussions from the persuasive online forum \emph{/r/ChangeMyView} (CMV). 
Specifically, we generate counter-arguments for CMV posts (see blue box in \Cref{fig:example}) using three LLMs and compare the likelihood of these texts expressing nine social dimensions to that of human-written opinion-changing comments.
We find that LLM-generated texts contain more social dimensions than human counter-arguments and consistently express \emph{trust} in the opinion holder (\S\ref{sec:results-social-dimensions-in-llms}). Examining the interaction dynamics, we observe that LLMs exhibit a capacity to model the communicative intent of the opinion holder and craft replies that mirror persuasive human strategies -- in the case of \gptthree, even more strongly than the opinion-changing human responses (\S\ref{sec:results-dynamics}). Finally, in a direct comparison between human-written opinion-changing comments and LLM-generated ones, crowdworkers found the AI-generated responses \emph{more agreeable} and overwhelmingly preferable. In fact, in 83\% of cases, they judged LLM-generated arguments as \emph{more likely to change the opinion holder's view} (\S\ref{sec:results-crowdsourcing}). 

Our work makes two key contributions. First, in the field of persuasive AI, we show that LLMs' persuasive capabilities extend beyond generating high-quality arguments. This underscores the need to study not just \emph{what} AI ``says'' but \emph{how} it engages with and adapts to human discourse, shaping opinion formation in new and unexpected ways. Second, regarding communication dynamics, we build on findings of \citet{monti2022language}, extending them to AI-driven interactions. Our results suggest that human preferences are influenced by more than just the quality of arguments, highlighting the complex interplay between social intent and persuasion.

\section{Related Work}
\label{sec:related_work}
\begin{table*}[th!]
    \centering
    \scriptsize
    \resizebox{\textwidth}{!}{
    \begin{tabular}{@{}l|l|l@{}}
        \toprule
        \textbf{Dimension} & \textbf{Description} & \textbf{Example}\\
        \midrule
        \textbf{Knowledge} & Exchange of ideas or information; learning, teaching & \emph{History shows the benefits of vaccinations outweigh the risks.}\\
        \textbf{Power} & Having power over behavior and outcomes of another & \emph{The only rights which exist in reality are legal rights.}\\
        \textbf{Status} & Conferring status, appreciation, gratitude, or admiration & \emph{I have nothing but the utmost respect for service men.}\\
        \textbf{Trust} & Will of relying on the actions or judgments of another & \emph{I understand your perspective.}\\
        \textbf{Support} & Giving emotional or practical aid and companionship & \emph{I’d feel sympathy for their situation.} \\
        \textbf{Similarity} & Shared interests, motivations or outlooks & \emph{I’m glad to know we agree on this.}\\
        \textbf{Identity} & Shared sense of belonging to the same group & \emph{They are members of their tribe (...)}\\
        \textbf{Fun} & Experiencing leisure, laughter, and joy & \emph{Realize how much you sound like Chamberlin when Hitler (...)}\\
        \textbf{Conflict} & Contrast or diverging views & \emph{Atheists are the worst for this (...)}\\
        \midrule
    \end{tabular}
    }
    \caption{Social dimensions of language, historically recognized in the social sciences and analyzed in a survey by \citet{deri2018SocialDims}. Examples are taken from the data and \citet{monti2022language}.}
    \label{tab:social-dims}
\end{table*}

Much of the literature on LLM persuasiveness is focused on \textbf{humans' perceptions of it}, often comparing LLMs to humans. \citet{karinshak2023Working} examined pro-vaccination messages created by both AI models and human authors, discovering that AI-generated messages were often perceived as more persuasive, except when they were explicitly labeled as AI-generated. Similarly, \citet{Goldstein2024How} found that GPT-3 could produce highly persuasive text, as measured by participants’ agreement with propaganda theses in a survey of U.S. respondents. \citet{bai_voelkel_eichstaedt_willer_2023} conducted a randomized controlled trial exposing a diverse group of individuals to policy commentaries written by either humans or LLMs and found that both methods were equally effective in influencing participants' policy support. \citet{hidden_potter_2024} showed that likely voters changed opinions and expressed a desire for more interaction after engaging with an LLM, even without being prompted to persuade them.

\paragraph{Personalization} plays a key role in the persuasiveness of models, as tailoring content to align with individuals' psychological traits can significantly influence their online behavior \cite{psychological_matz_2017, review_teeny_2020}. \citet{potential_matz_2024} demonstrated that personalized messages crafted by ChatGPT were significantly more persuasive than non-personalized ones. \citet{measuring_pauli_2024} found that assigning different personas to LLMs could significantly alter the text's persuasiveness. Similarly, \citet{durably_costello_2024} engaged conspiracy believers in personalized, evidence-based dialogues with GPT-4-turbo and reported a reduction in conspiracy beliefs by 20\%. In a controlled study by \citet{salvi2024conversationalpersuasivenesslargelanguage}, concerns were raised about the implications of personalized persuasion for the governance of social media and online environments. Lastly, \citet{sharma2024towards} analyzed human preference data showing that when AI responses matched users’ views, they were more likely to be preferred. Their research also suggested that both humans and preference models sometimes favored sycophantic responses over correct ones, highlighting a potential issue in AI behavior.

While a small body of research has explored whether LLMs are persuasive to humans, there is a lack of literature explaining \textbf{\emph{why} they are persuasive, particularly in relation to established theoretical frameworks}. Closely related to our work, \citet{persuasive_breum_2024} examined the effectiveness of LLM-generated arguments when designed to convey specific social dimensions. Their findings showed that arguments combining factual knowledge, trust markers, expressions of support, and status cues were rated as most persuasive by both humans and AI agents. Notably, humans found knowledge-based arguments particularly compelling, highlighting the importance of factual support in persuasive messaging. However, their study was conducted in a synthetic persuasion dialogue setting, leaving open the question of whether LLMs naturally incorporate social dimensions when countering real-world social media posts without being explicitly prompted. Furthermore, no direct comparison has been made between LLM-generated and human-written messages regarding the presence of social dimensions or their effects on human perceptions of persuasiveness.

\section{Data}
\label{sec:data}
This study draws on discussions from the subreddit \emph{/r/ChangeMyView}, a forum where users share their opinions and invite others to challenge them. When a commenter successfully persuades the original poster (OP) to reconsider their stance, they receive a ``delta'' ($\Delta$) as a reward, signaling a successful change of opinion (see $\Delta$ example in \Cref{fig:example}).

We use the publicly available ChangeMyView (\textbf{CMV}) corpus collected by \citet{Tan2016CMV}. It is a well-established dataset that underwent multiple further analyses \cite[inter alia]{hidey-etal-2017-analyzing,falenska-etal-2024-self}. Crucially, since the dataset consists of posts and comments written before 2016, we can be certain that none were generated by LLMs. Since we do not train models on this data, we merge the training and held-out portions for analysis, which consists of $20,626$ posts and $1,260,266$ comments.

\subsection{Filtering Posts and Comments}
\label{subsec:data-filtering}
The raw CMV includes deleted posts and Reddit-specific meta-level comments. To refine the data, we removed entries with missing or deleted text, missing corresponding posts or comments, or were too short to present a meaningful argument (\small{$\leq$} \normalsize$2$ words, whitespace-tokenized). After this cleaning process, we were left with $20,151$ posts and $1,193,483$ comments.

\subsection{Extracting Sociopolitical Posts}
\label{subsec:sociopolitical-classifier}
For compatibility with \citet{monti2022language}, we use only sociopolitical posts in our analysis. For filtering, we use their classifier trained on data from sociopolitical subreddits (detailed in \Cref{app:sociopolitical-classifier}), obtaining $13,504$ posts and $864,890$ comments. Henceforth, this is the data used in the experiments.

\subsection{Finding Delta Comments}
\label{subsec:delta-annotations}
\emph{/r/ChangeMyView} forum employs a DeltaBot to allow OPs to award $\Delta$s to opinion-changing comments. Whenever an OP awards a $\Delta$, the bot generates a comment within the same thread, marking the user who received it.
Since these meta-level comments can appear at any point in the thread -- when the OP decides that the author has changed their view, which may have occurred several messages earlier -- we treat all parent comments by the author within the thread as the opinion-changing ones. We refer to these as \textbf{$\Delta$ comments}, totaling $7,254$ in our dataset.

\section{Methods}
\label{sec:methodology}
\subsection{LLMs and Counter-Argument Generation}

Our study investigates \textbf{whether} state-of-the-art LLMs convey illocutionary intent in argumentative contexts. To enable a focused and in-depth analysis, we select three representative models -- chosen for their widespread use and strong performance in generating high-quality arguments -- since these are the most likely to express meaningful illocutionary intents and carry societal relevance.

\paragraph{LLMs}
\label{subsec:models}
We start from a selection of seven widely used LLMs, including five open-access models: \textbf{\llamatwo (\llamatwosevenOnlyNr, \llamatwoseventyOnlyNr}, \citet{Touvron2023Llama2}), \textbf{\llamathree (\llamathreeeightOnlyNr, \llamathreeseventyOnlyNr}, \citet{Touvron2023Llama2}), and \textbf{\mistral} \citep{jiang2023mistral}, and two API-access \gpt models: \textbf{\gptthree} \cite{Brown2020GPT3} and \textbf{\gptfour} \citep{openai2023gpt4}.\footnote{Accessed between Dec. 8 - 19. 2024.}

\paragraph{Inference and prompts}
\label{subsec:inference}
We use the HuggingFace text generation inference pipeline\footnote{\url{https://huggingface.co/docs/text-generation-inference}} for open-access and the OpenAI text completion API\footnote{\url{https://platform.openai.com/docs/guides/gpt}} for the API-access models. 
To simulate an interaction in which an LLM attempts to change an individual's opinion, we take an original post and append \emph{``You have one chance to change my view. Answer:''}. We then prompt the models with nucleus sample decoding (temp$=$$.9$, top\_p$=$$.6$) and obtain a single counter-argument per post from each model. All generated outputs are capped at $600$ tokens.

\paragraph{LLM selection}
To select the final three models for the analysis, we sub-sampled $50$ sociopolitical posts from CMV and generated counter-arguments to them with all seven models. We then examined their outputs with the argument quality classifiers from 
\citet{falk-lapesa-2023-bridging}.
Based on the \emph{effectiveness} and \emph{impact} scores -- two measures that quantify the persuasiveness and the degree of likability of the arguments -- we selected three final models: \textbf{\llamatwoseven}, \textbf{\mistral}, and \textbf{\gptthree}. 
Since this analysis is not the focus of our work, we leave the details for the \Cref{par:arg-quality-evaluation}. In general, 
%we found that 
the \llamathree family generated substantially lower quality arguments compared to the rest. Secondly, \mistral overall scored the highest. % among all models. 
Finally, \llamatwo and \gpt models generated arguments that are similar in quality within their families. The scores among \llamatwo models were not significantly different (t-test with $\alpha<0.01$). Thus, we chose the smaller model in favor of reducing the environmental and computational costs. Similarly, \gptfour did not generate significantly dissimilar arguments than \gptthree. Since it is approx. fifteen times more expensive and currently suffers from API-request timeouts due to high demand, we selected the smaller model.

\subsection{Measuring Social Dimensions}
To infer and compare the illocutionary intents conveyed in CMV posts, comments, and LLM-generated counter-arguments, we ground the analysis in the theoretical model of social dimensions (see examples in \Cref{tab:social-dims}). Throughout this process, we directly apply the methodology from \citet{monti2022language} and below explain the required deviations.

\paragraph{Social dimension extraction}
To extract social dimensions from texts, we use classifiers developed by \citet{choi2020TenDims}. The models estimate the likelihood that a text $t$ conveys a social dimension $d$ by predicting a score between $0$-$1$ (i.e., least likely to most likely). For each text, we obtain scores for nine dimensions from their respective models and then binarize them with a $85^{th}$ percentile threshold.

\citet{monti2022language} reports that the probability of being labeled with $d$ naturally increases with the length of the message. Thus, not to penalize the short arguments, we follow their approach and apply weight discounting to the binary dimension scores (for more details, see \Cref{app:social-dimensions}).

\paragraph{Social intent and opinion change}
To analyze the relationship between social dimensions and opinion change (indicated by the presence of $\Delta$) across various texts, we follow \citet{monti2022language} and use \textbf{the odds ratios (OR)} as the measure of the association of a dimension $d$ (further details are available in \Cref{app:odds-ratios}):

\begin{equation}
    \text{OR}(p(d|\Delta), p(d|\overline{\Delta})) = \frac{\text{odds}(p(d|\Delta))}{\text{odds}(p(d|\overline{\Delta}))},
\label{eq:odds-ratios}
\end{equation}

\noindent where $\text{odds} = \frac{p}{1 - p}$, $\overline{\Delta}$ are comments that did not receive $\Delta$, and $p(d|\Delta)$ is the conditional probability that a comment contains $d$, given that it received $\Delta$ (similarly for $\overline{\Delta}$). 

To compare human-written and LLM-generated comments, we use a comparable odds ratio expression: $\text{OR}(p(d|\text{LLM}), p(d|\Delta))$, where we contrast the likelihood of a dimension $d$ appearing in LLM-generated comments versus in human-authored $\Delta$ comments.

Finally, to assess the impact of illocutionary intent reciprocation on opinion change, and to determine whether LLMs exhibit patterns similar to those of successful human commenters (see \textbf{RQ3}), we measure the odds ratios for post-comment pairs (analogously for $\overline{\Delta}$ and LLM generated texts):

\begin{equation}
    \text{OR}(p_\Delta(d_i|d_j), p(d))\\
    = \frac{\text{odds}(p_\Delta(d_i|d_j))}{\text{odds}(p(d))},
\label{eq:odds-ratios-cond}
\end{equation}

\noindent where $p_\Delta(d_i|d_j)$ is the conditional probability of a $\Delta$ comment containing dimension $d_i$, given that its corresponding post contains dimension $d_j$, and $p(d)$ is the prior probability of a message being labeled with dimensions $d$. We use $p(d)$ as an offset representing the baseline probability that a post with a dimension $d_j$ receives a comment with dimension $d_i$ at random -- essentially estimating the chance of that dimension occurring in a message at random. 
We report the variation of the probability of achieving $\Delta$ given a combination of dimensions (analogously to compare $p(d_i|d_j)$ between the LLM-generated and $\Delta$ comments) only for the statistically highly significant odds ratio variations (P\ $< 0.01$) calculated as $\text{OR}(p_\Delta(d_i|d_j), p(d)) - \text{OR}(p_{\overline{\Delta}}(d_i|d_j), p(d))$.

\section{Results}
\label{analysis:results}
\begin{figure}[t]

    \begin{subfigure}[t]{\columnwidth}
        \centering
        \includegraphics[width=0.89\columnwidth, trim=0 0 0 0, clip]{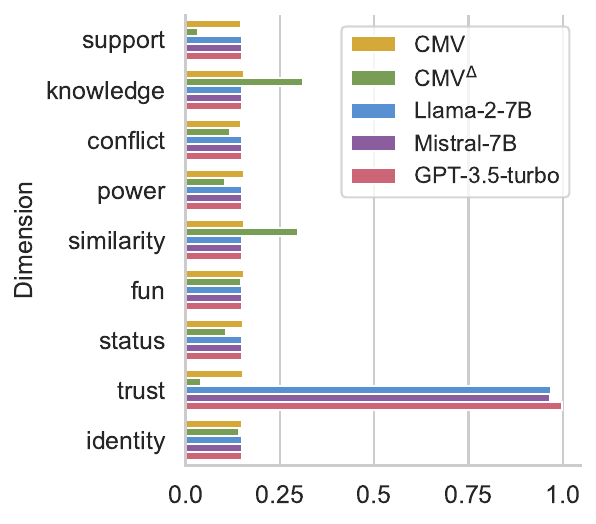}
        \caption{\% of comments containing a particular dimension}
         \label{fig:dimension-freqs}
    \end{subfigure}%
    
    \vspace{0.5cm}
   
    \begin{subfigure}[t]{\columnwidth}
        \centering
        \includegraphics[width=0.65\columnwidth, trim=0 0 0 0, clip]{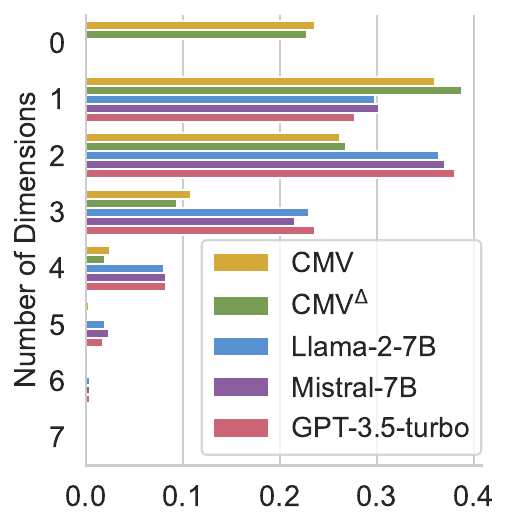}
        \caption{\% of comments containing a certain number of dimensions}
        \label{fig:num-dimensions}
    \end{subfigure}

    \caption{\label{fig:comm_stats} Statistics of comments written by CMV users and generated by LLMs.}
\end{figure}

\begin{figure*}[t]

    \begin{subfigure}[t]{0.25\textwidth}
        \centering
        \includegraphics[height=2.1in]{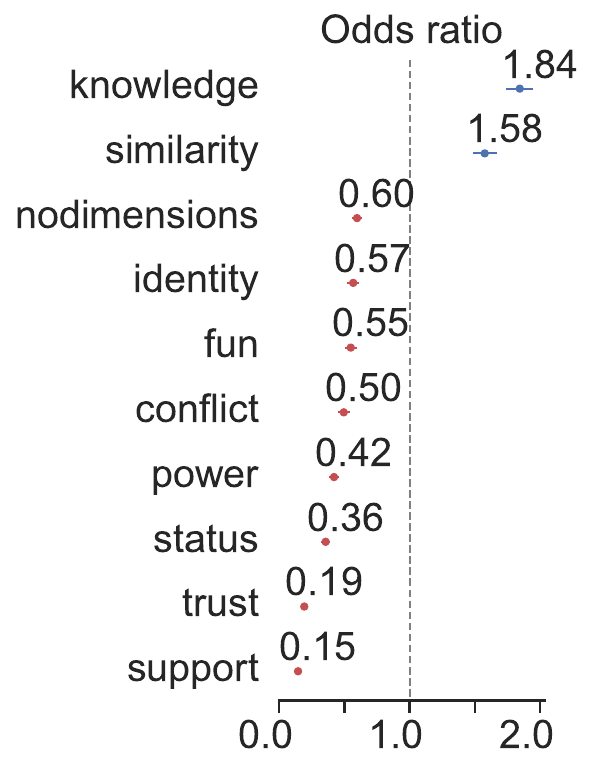}
        \caption{$\Delta$ vs. $\overline{\Delta}$}
         \label{fig:human-odds}
    \end{subfigure}%
    \begin{subfigure}[t]{0.25\textwidth}
        \centering
        \includegraphics[height=2.1in]{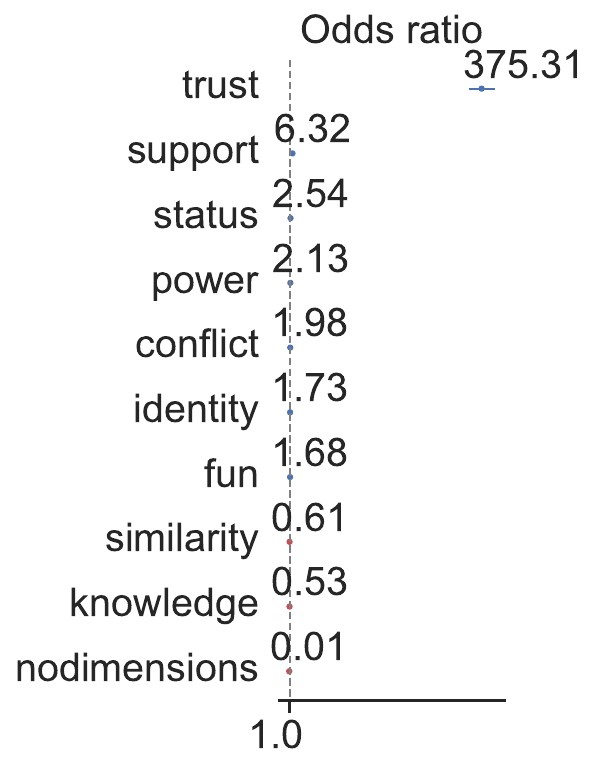}
        \caption{\label{fig:odds-llama} \llamatwoseven vs. $\Delta$}
    \end{subfigure}%
    \begin{subfigure}[t]{0.25\textwidth}
        \centering
        \includegraphics[height=2.1in]{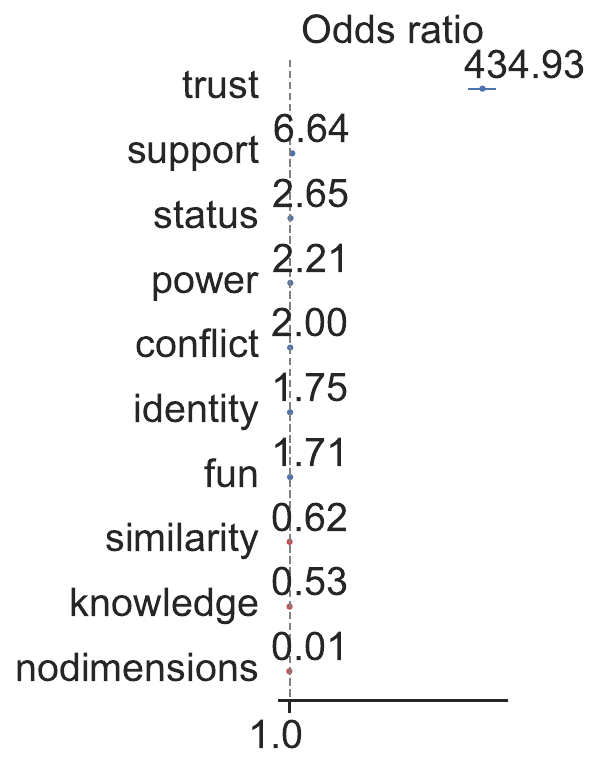}
        \caption{\label{fig:odds-mistral} \textsc{Mistral}-7B vs. $\Delta$}
    \end{subfigure}%
    \begin{subfigure}[t]{0.25\textwidth}
        \centering
        \includegraphics[height=2.1in]{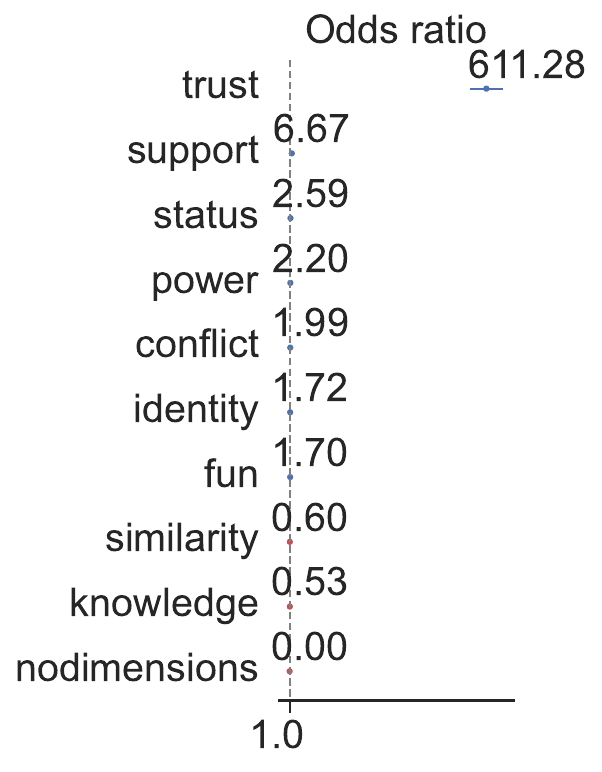}
        \caption{\label{fig:odds-gpt} \gptthree vs. $\Delta$}
    \end{subfigure}
    \caption{Odds ratios between probabilities of social dimensions present in texts measured as in \Cref{eq:odds-ratios}. (a) comparison of human-written $\Delta$ and $\overline{\Delta}$ comments; (b), (c), (d) comparison of LLM-generated and $\Delta$ comments. Error bars represent the 95\% confidence intervals (see \Cref{eq:confidence-interval} and details in \Cref{app:odds-ratios}).}
    \label{fig:odds-ratios}
\end{figure*}

In the following,  we analyze social dimensions in counter-argument comments generated by three selected models: \llamatwoseven, \mistral, and \gptthree. 

\subsection{Social Dimensions in LLM Comments}
\label{sec:results-social-dimensions-in-llms}
We start by addressing \textbf{RQ1}: \textit{Which social dimensions are present in LLM-generated arguments?}
\Cref{fig:dimension-freqs} illustrates the percentage of messages containing a given dimension across different sets. First, human-written comments (yellow bars) show relatively balanced distributions across dimensions, indicating that CMV users do not favor any particular social dimension-based rhetorical strategy. However, the $\Delta$ comments (green bars) more frequently contain \emph{knowledge} and \emph{similarity} -- a result that corroborates \cites{monti2022language} findings. 

In contrast, all three LLMs overwhelmingly express \emph{trust} -- often starting with phrases like \textit{I understand your perspective} -- with nearly all generated messages incorporating this dimension. This finding sharply differentiates LLM-generated texts from $\Delta$ comments, where \emph{trust} is the second least frequent dimension. Beyond overusing \emph{trust}, LLMs also produce responses that incorporate more social dimensions within a single message. \Cref{fig:num-dimensions} shows that most human comments (yellow and green bars) tend to rely on a single dimension. Moreover, around $23$\% contains no identifiable dimensions, while only about $2$\% includes four. In contrast, generated messages consistently include at least one dimension, with some containing as many as six. 

In summary, models not only heavily rely on markers of \emph{trust} but also convey consistently \emph{more social intent} than humans -- a pattern that remains remarkably consistent across all three LLMs.

\subsection{LLMs vs. Opinion-changing Comments}
\label{sec:results-odds-model-vs-delta}
Having confirmed the presence of social dimensions in LLM-generated arguments, we explore \textit{further differences between these texts and human-written opinion-changing comments} (\textbf{RQ2}).

\paragraph{$\Delta$ vs. $\overline{\Delta}$ comments}
We start by identifying the social dimensions vital for opinion-changing human arguments. \Cref{fig:human-odds} presents odds ratios (see \Cref{eq:odds-ratios}) between the $\Delta$ and $\overline{\Delta}$ comments. We observe that $\Delta$ comments are more likely to convey \emph{knowledge} ($84$\%)\footnote{A ratio of $1$ shows no difference, while a ratio of $1.84$ reads as \emph{$84$\% more likely}.} and \emph{similarity} ($58$\%) compared to $\overline{\Delta}$. Conversely, comments that do not contain any dimension (nodimensions) are $40$\% less likely to change the opinion holder's view.\footnote{The differences from \citet{monti2022language} likely arise from a different selection of CMV posts, yet the main findings remain consistent.}

\paragraph{LLMs vs. $\Delta$ comments}
We shift the focus back to LLMs and compare generated arguments with $\Delta$ comments in \Cref{fig:odds-llama,fig:odds-mistral,fig:odds-gpt}. The odds of dimensions present in model-generated messages across all three models are highly similar. Consistent with the findings from the previous section (see \emph{trust} in \Cref{fig:comm_stats}), all three LLMs are far more likely to express \emph{trust} than $\Delta$ comments. Similarly, models are highly unlikely to generate arguments that do not convey any social dimension. Compared to the $\Delta$ comments, models are more likely to convey \emph{support}, \emph{status}, \emph{power}, \emph{conflict}, \emph{identity}, and \emph{fun} by approx. $554$\%, $159$\%, $118$\%, $99$\%, $73$\%, $70$\% respectively. Conversely, they are less likely to convey \emph{similarity} and \emph{knowledge} by $39$\% and $47$\%.

In conclusion, LLM-generated comments convey more illocutionary intent within a single comment than human-written ones and exhibit similar distributions of social dimensions, with the exception of \emph{trust}. However, they differ markedly from successful ($\Delta$) human-written comments. In particular, LLMs are less likely to express the two dimensions most strongly associated with opinion change: \emph{knowledge} and \emph{similarity}. 

\begin{figure}[t!]
    \centering
    \includegraphics[height=2.5in, trim=0 0 0 0, clip]{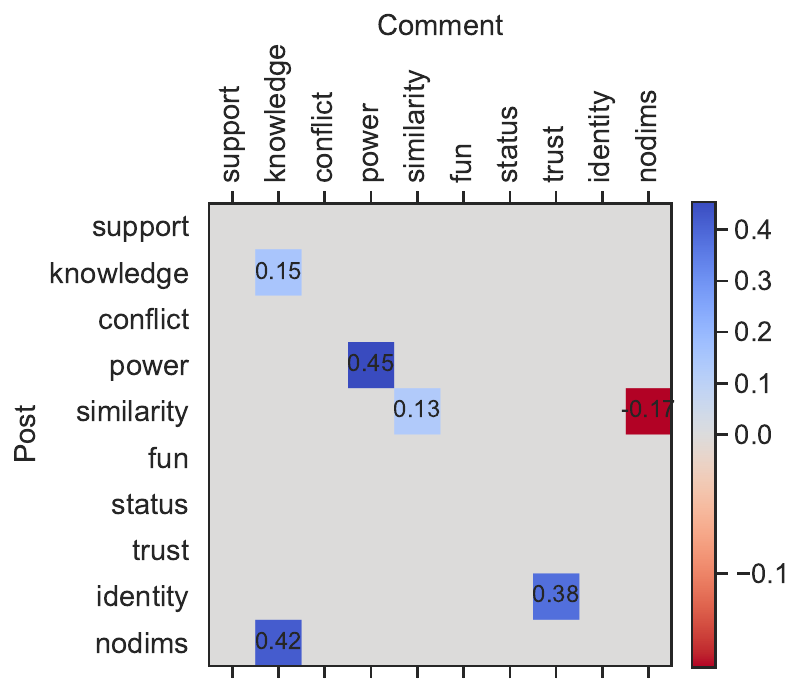}
    
    \caption{Variation of the probability of expressing a particular combination of dimensions in post-comment pairs measured by taking the difference of odds ratios calculated in \Cref{eq:odds-ratios-cond} between $\Delta$ and $\overline{\Delta}$ comments.}
    \label{fig:odds-cond-human}
\end{figure}

\subsection{Social Intent Dynamics}
\label{sec:results-dynamics} 
\textit{To what extent can LLMs capture interactional dynamics of social dimensions between opinion holders and successful commenters?} (\textbf{RQ3})? To answer this question, we first establish the dynamics in human-human conversations. We then contrast characteristics of LLM-generated comments with human-written opinion-changing comments.

\paragraph{Human-human conversations}
\Cref{fig:odds-cond-human} shows the differences between the $\Delta$ and $\overline{\Delta}$ post-comment pairs measured by taking the difference of odds ratios calculated in \Cref{eq:odds-ratios-cond}.
We observe a similar phenomenon as reported by \citet{monti2022language}, i.e., comments that make similar appeals to that of their corresponding posts are more likely to receive a $\Delta$ % from the OP 
(blue cells on the diagonal in the figure). We observe this pattern for three out of nine dimensions: if a comment conveys \emph{knowledge}, appeals to \emph{power} or \emph{similarity}, it is more likely to receive a $\Delta$ by $15$\%, $45$\%, and $13$\% resp. Moreover, if a comment expresses \emph{trust} in response to a post appealing to \emph{identity} or \emph{knowledge} in response to no dimensions, it is more likely to be awarded a $\Delta$ by $38$\% and $42$\%, resp. Conversely, if a post appeals to \emph{similarity}, a comment containing no dimensions is $17$\% less effective.

\begin{figure*}[t]
    \begin{subfigure}[t]{0.34\textwidth}
        \centering
        \includegraphics[height=2.45in, trim=0 0 2cm 0, clip]{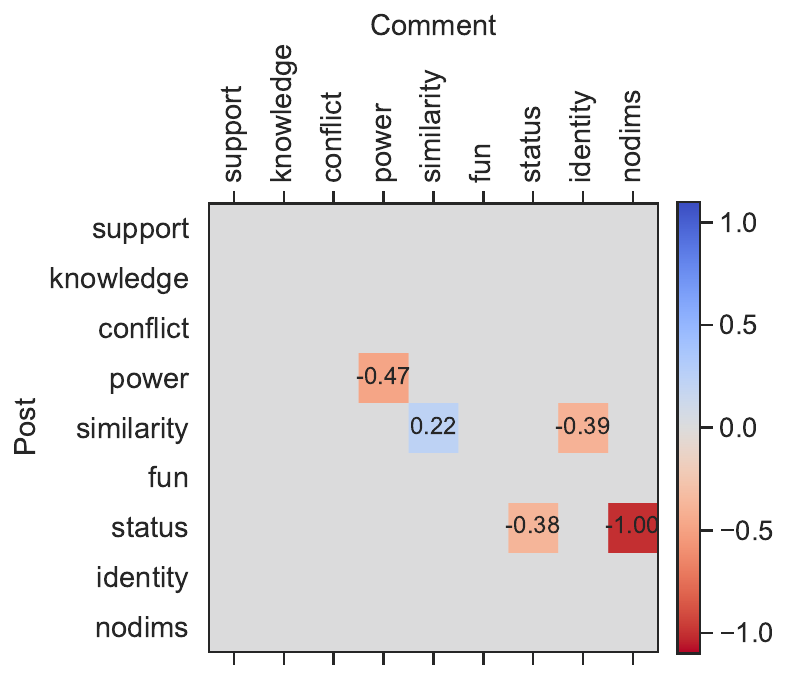}
        \caption{ \label{fig:llama-cond} \llamatwoseven vs. $\Delta$}
    \end{subfigure}%
    \hfill
    \begin{subfigure}[t]{0.28\textwidth}
        \centering
        \includegraphics[height=2.45in, trim=3.2cm 0 2.05cm 0, clip]{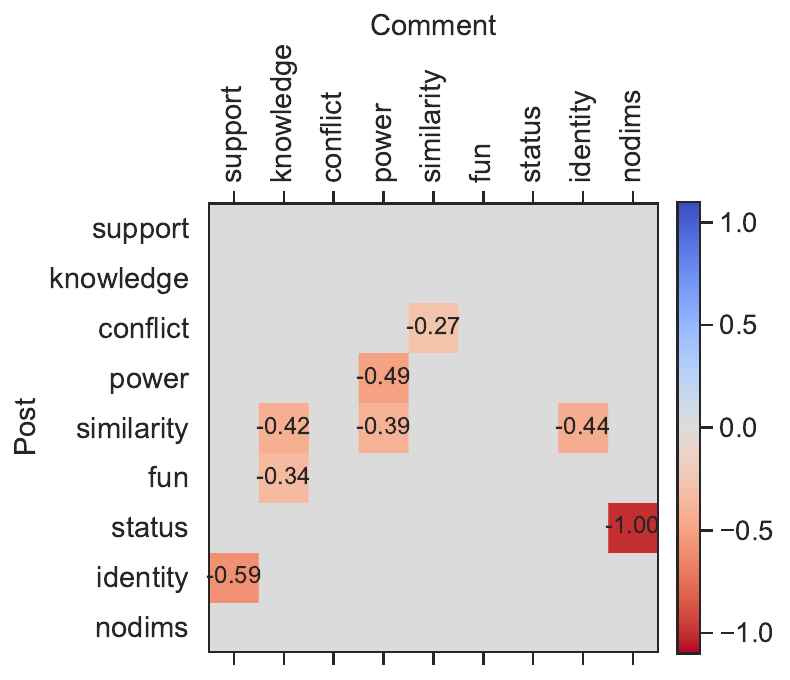}
        \caption{ \label{fig:mistral-cond} \textsc{Mistral}-7B vs. $\Delta$}
    \end{subfigure}%
    \begin{subfigure}[t]{0.34\textwidth}
        \centering
        \includegraphics[height=2.45in, trim=3.2cm 0 0.1cm 0, clip]{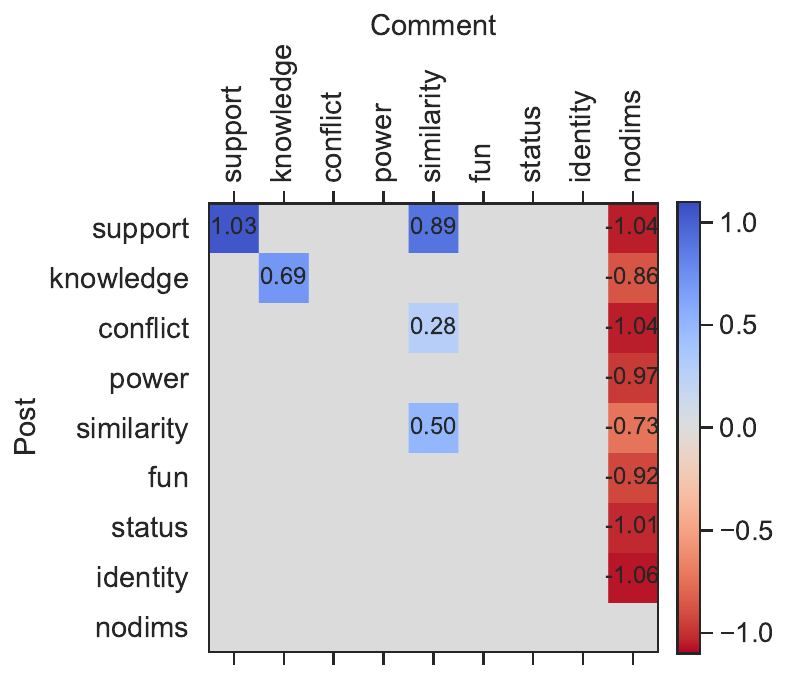}
        \caption{ \label{fig:gpt-cond} \gptthree vs. $\Delta$}
    \end{subfigure}
    \caption{Variation of the probability of expressing a particular combination of dimensions in post-comment pairs measured by taking the difference of odds ratios calculated in \Cref{eq:odds-ratios-cond} between LLM-generated and human-written $\Delta$ comments, displayed for statistically highly significant variations (P\ $< 0.01$).}
    \label{fig:odds-cond-llm}
\end{figure*}

\paragraph{LLM-generated arguments}
\Cref{fig:odds-cond-llm} presents the variation of the probability of expressing a particular combination of dimensions between LLM-generated texts and $\Delta$ comments (note that the \emph{trust} dimension is removed due to its high occurrence and to prevent visual clutter). The highlighted areas show where model-generated comments significantly differ from the $\Delta$ comments: in five dimension pairs for \llamatwoseven, eight for \mistral, and thirteen for \gptthree. For \llamatwoseven, the odds of the model conveying \emph{similarity} in its reply to a post with the same dimension is $22$\% more likely than the CMV $\Delta$ comments. The opposite happens for \emph{power} where the model is less likely to make a similar appeal by $47$\%. 
Interestingly, \mistral, the only non-chat model in this study, is the most dissimilar to human social dimension dynamics. The model is overall less likely to reply with certain dimensions than the $\Delta$ comments in $8$ out of $81$ pairs of dimensions.

Focusing on the diagonal -- i.e., whether comments reciprocate the social dimensions expressed in the original posts -- we find that model-generated comments exhibit patterns similar to those in successful ($\Delta$) human-written comments. For \llamatwoseven and \gptthree, there is no statistically significant difference in six out of nine cases, and for \mistral, in eight out of nine. On average, \llamatwoseven and \mistral tend to under-reciprocate compared to $\Delta$ comments, whereas \gptthree shows a notable deviation: it is significantly more likely to reciprocate social dimensions -- \emph{support} (by 103\%), \emph{knowledge} (69\%), and \emph{similarity} (50\%) -- than human $\Delta$ comments. Furthermore, \gptthree is also more likely to appeal to \emph{similarity} in response to \emph{support} (by 89\%) and \emph{conflict} (by 28\%).

To sum up, models not only differ from $\Delta$ comments but also from each other when it comes to the post-comment dynamics of social dimensions. Among them, \gptthree exhibits stronger reciprocity by aligning more closely with the post's intent. This suggests that if individuals are, in fact, more likely to change their view when a challenger demonstrates similar illocutionary intent in their counter-argument, then they might be similarly (or even more) likely to be convinced by GPT-generated arguments.

\subsection{Human Preferences}
\label{sec:results-crowdsourcing}
We observed significant differences in how CMV users and LLMs convey illocutionary intent. \textit{Do these differences correlate with human preferences} (\textbf{RQ4})? To answer this question, we focus on the two differences that significantly stood out throughout our analyses so far: the overwhelming presence of the \emph{trust} dimension in the LLM-generated texts and the comparably lower use of \emph{knowledge} -- a dimension strongly associated with opinion change.

\paragraph{Method}
Since it is not possible to ask the original CMV posters if they would change their opinion at the time when they wrote their posts, we approximate this situation through crowdsourcing. In an exploratory study, we showed human annotators an original CMV post and first asked whether they agreed with its stance. Then, they were presented with two counter-arguments -- one human-written $\Delta$ comment and one LLM-generated -- and asked whether they agreed with any of them. Annotators could indicate agreement or disagreement with a post or a comment independently of their agreement with the other. Finally, they were asked to \emph{select the message more likely to change the opinion holder's view}.

We annotated 100 triples $\langle$CMV post, $\Delta$ comment, LLM comment$\rangle$, with each reviewed by three annotators. The triples were sampled to equally represent \emph{trust} and \emph{knowledge} dimensions (25 for each combination from \Cref{tab:agreement}). Finally, since \gptthree stands out most in its expression of social intent dynamics (see \S\ref{sec:results-dynamics}) and is among the most widely used language models, we selected arguments generated by this model for our study.

Annotators were given guidelines, including a tutorial on how to perform the task. The annotation was performed on prolific\footnote{\url{https://www.prolific.com/}}, and the worker pool consisted of U.S. residents\footnote{Crowdworkers were fluent English speakers aged 18 or older. They represented diverse ethnicities and countries of birth, and the sample was gender-balanced between female and male -- the only gender variables available on the platform.}. Each annotator annotated eight instances and got compensated £12/h (more details on crowdsourcing in \Cref{sec:appendix_b}).

\paragraph{Results}
Overall, crowdworkers found \gptthree-generated messages more persuasive, selecting them as \emph{more likely to change the opinion holder’s view} in $83$\% of cases according to the majority vote. The agreement among the annotators was measured at $0.79$ using Krippendorff’s alpha.

\begin{table}[t]
    \centering
    \scriptsize
    \resizebox{0.5\textwidth}{!}{
    \begin{tabular}{@{}l|l|l|l@{}}
    \multicolumn{2}{c}{} & \multicolumn{2}{c}{\textbf{ $\Delta$ Comments}} \\
        \cmidrule(lr){3-4}
    \multicolumn{1}{c}{} & \multicolumn{1}{c|}{\textbf{Dimension}} & \multicolumn{1}{l}{Knowledge} & \multicolumn{1}{l}{Trust}\\
        \cmidrule(lr){2-4}
        \textbf{\gpt-} & Trust & 0.62 & 0.85\\
        \textbf{Generated} & Knowledge and Trust & 0.81 & 1.00\\
    \end{tabular}
    }
    \caption{Krippendorf's alpha scores as inter-annotator agreement obtained from three binary preference ratings per dimension pairs ($25$ pairs of a $\Delta$ from CMV and \gptthree-generated comment per category).}
    \label{tab:agreement}
\end{table}

We further analyzed these results from two angles. First, we found that the preference for opinion-changing arguments was not affected by the combination of \emph{knowledge} and \emph{trust}. That is, whether both the human and model responses relied solely on \emph{trust}, or the model’s \emph{trust}-based argument was compared to a human argument grounded in \emph{knowledge}, the preference ratio remained largely consistent. Additionally, we observed a higher Krippendorff’s alpha for samples where model responses included both \emph{knowledge} and \emph{trust}, compared to those based on \emph{trust} alone, which indicates stronger agreement among annotators (see \Cref{tab:agreement}).

Second, we examined the relationship between annotators' opinions on the comment stances and their preferences. In 58 cases, the crowdworkers indicated a (dis-)agreement with both counter-arguments while still, on average, preferring generated ones. In the majority of the remaining cases (37 out of 42 posts), annotators selected the counter-argument they also agreed with, suggesting a potential preference bias. Interestingly, in 84\% of these cases (31 out of 37), this selected and preferred comment was LLM-generated, implying that the crowdworkers found GPT-generated messages also \emph{more agreeable} (majority agreement with $92$ model comments and $66$ human $\Delta$ comments, see \Cref{tab:stance-agreement}).

\section{Conclusions and Discussion}
\label{sec:conclusion}
As LLM-generated content becomes increasingly prevalent online, understanding its influence on human opinions becomes essential. 
To move towards this understanding, we examined LLM persuasiveness through the lens of Communicative Action -- a theoretical framework that views language as a means of expressing social intent through reasoned dialogue aimed at achieving mutual understanding, which is crucial for human opinion change.

Using real-world persuasive discourse as a reference point, we showed that models effectively convey illocutionary intent -- often more frequently and densely than humans. Notably, all three models we analyzed consistently express \emph{trust} in their conversation partners, potentially affirming their views and contributing to perceived biased likability \cite{sharma2024towards}. 
While all three models show similar patterns of social intent reciprocity to that of the successful human comments, the post-comment interactions of \gptthree-generated arguments reveal even stronger patterns of reciprocity -- a behavior closely linked to opinion change \cite{monti2022language}.

Finally, LLMs employ rhetorical strategies that differ from those found in human-written opinion-changing arguments -- especially in conveying the two dimensions previously linked to successful persuasion: \emph{knowledge} and \emph{similarity}. Importantly, crowdworkers not only find these LLM-generated arguments more agreeable but also consistently prefer them over human-written opinion-changing ones. While further research is needed to identify which dimensions drive their decision the most, our findings show that LLMs are capable of engaging in \emph{nuanced communicative actions}, potentially shaping human opinions in ways not yet fully understood. Moreover, the varying patterns observed between instruction and chat models suggests that alignment training for LLMs may not only reinforce existing human biases but also increase individuals’ susceptibility to AI-driven influence on critical issues like politics, social justice, and the environment. In light of these concerns, we urge further research into alignment training and its broader effects on public discourse and opinion formation.

\section{Limitations}
\label{sec:limitations}
We simulate exchanges between opinion holders and LLMs using data from the \emph{/r/ChangeMyView} subreddit. Discussions on other online platforms and real one-to-one conversations with LLMs might have different characteristics,  highlighting an open area for further research.

Additionally, this study has some constraints related to the exploratory annotation experiment. First, humans write texts with various lengths, some substantially shorter or longer than the average. In comparison, the artificial setup in which the LLM-generated texts have a fixed maximum length might make them more or less expressive than human-written comments. Although this does not affect our findings in the main study, as we normalize probabilities by length, it might affect human reading comprehension and perception. In other words, arguments might have different writing styles depending on the length, potentially influencing human preferences. To account for these effects, we sampled both human-written and LLM-generated comments with varying lengths.

Second, we do not control for potential topic confounders in the annotation study. Discussion topics may influence both the model-generated texts and the crowdworkers' judgments, as annotators are biased in agreeing with messages that align with their prior beliefs. Our randomly sampled annotation instances contain various topics to account for this. To account for this, our annotation instances were randomly sampled from a variety of topics. However, the impact of topic confounders on opinion change remains an open area for future research.

Finally, in our annotation study, we ask crowd-sourced workers whether they believe a given argument could change an individual's opinion. Since these are not their own posts, they can only speculate on the argument’s persuasiveness. To mitigate this limitation, we also ask annotators to evaluate their own agreement with the post and comments and include this dimension in the analysis.

\section{Ethical Considerations}
\label{sec:ethical_considerations}
Understanding why LLMs are persuasive to humans has the potential dual-use risk. The developers of such models can use our findings to train LLMs to be more persuasive to specific target groups for harmful purposes. We not only \textbf{advise against such efforts} but also advocate for detecting and preventing manipulative model deployment.

\section{Acknowledgements}
We acknowledge the support of the Ministerium für Wissenschaft, Forschung und Kunst BadenWürttemberg (MWK, Ministry of Science, Research and the Arts Baden-Württemberg under Az. 33-7533-9 19/54/5) in Künstliche Intelligenz \& Gesellschaft: Reflecting Intelligent Systems for Diversity, Demography and Democracy (IRIS3D) and the support by the Interchange Forum for Reflecting on Intelligent Systems (IRIS) at the University of Stuttgart. We would like to thank Eva Maria Vecchi for her valuable feedback during the early stages of this work, and Neele Falk for her help in selecting appropriate metrics for evaluating argument quality.

\bibliography{anthology,custom}

\appendix

\section{Data and Methods}
\label{sec:appendix_a}
\subsection{Data}
\label{app:data}
This section provides additional details about the CMV forum and the filtering process for sociopolitical posts.

\subsubsection{CMV Forum}
\label{app:data-cmv-forum}
The subreddit CMV functions similarly to other Reddit forums. Users initiate discussions by posting a viewpoint in the title and elaborating on it in the body text, which may include external links to websites, images, or other resources. The platform is designed for opinion exchange: once a post is live, any Reddit user can engage with it by voting or commenting (a Reddit account is required to do so). Users can also reply to comments, forming a tree-structured discussion. A distinctive feature of CMV is the ability for the original poster to award a $\Delta$ flag to comments that successfully changed their view.

Note that, due to the nature of online discourse, the CMV dataset may contain offensive content. However, this content was not created by us, nor do we employ any methods that endorse or promote such material.

\subsubsection{Sociopolitical Classifier}
\label{app:sociopolitical-classifier}
The classifier (\{1,2,3\}-gram logistic regression model) has been trained on $10.000$ posts -- $50$ posts from each $51$ sociopolitical subreddit and later filtered between the years from $2011$ to the end of $2019$ -- and validated on an equal size test set with an average performance of $ 89.5\%$ (more details in \citet{monti2022language}).

\subsection{Methods}
This section provides additional details on model selection, LLM-based counter-argument generation, social dimension classification, and the metrics.
\subsubsection{LLM Selection}
\label{app:llm-selection}
\paragraph{Argument quality evaluation}
\label{par:arg-quality-evaluation}
\citet{falk-lapesa-2023-bridging} use RoBERTa 
(roberta-base, \citet{liu2019robertarobustlyoptimizedbert}) as the backbone transformer and train an adapter per dimension ($20$ single-task adapters). Using these adapters, we obtain scores for seven dimensions relevant to our purposes shown in \Cref{tab:arg-quality-dims}. These scores help us assess the quality of the generated arguments across models. 

\begin{figure*}[th!]
    \centering
    \resizebox{1\textwidth}{!}{
        \includegraphics{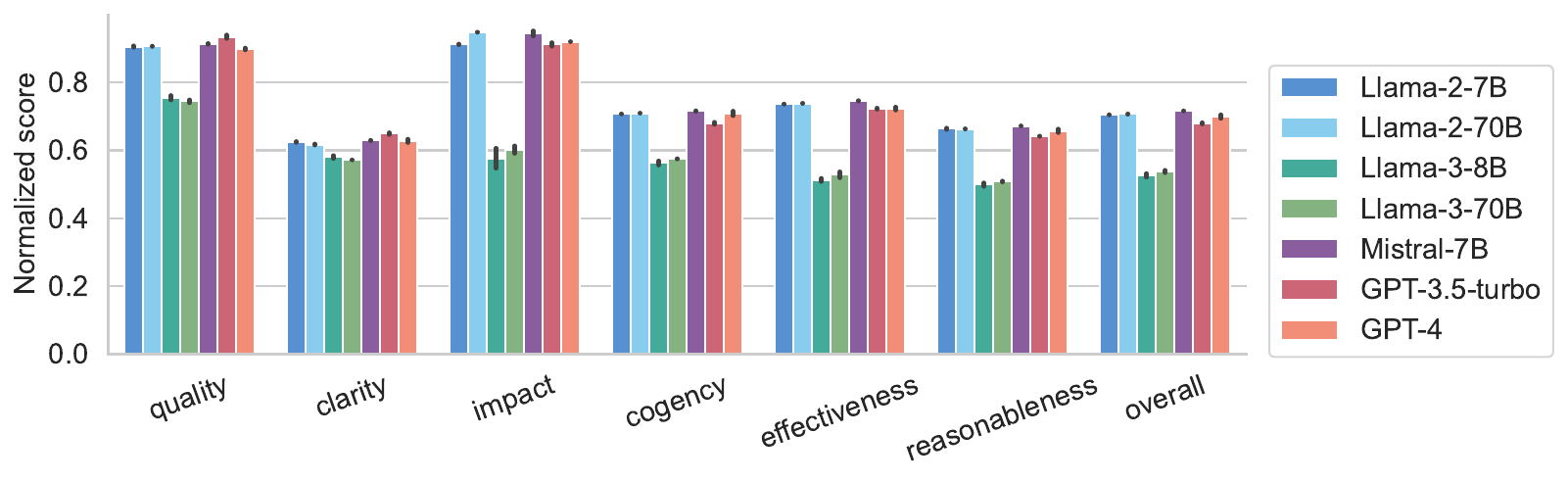}
    }
    \caption{Argument quality results from all seven models on CMV$^{50}$, normalized to range $(0,1)$ and averaged between the two inference settings (nucleus sampling and greedy decoding) with black bars representing the variance in scores between the decoding strategies.}
    \label{fig:arg-quality-initial}
\end{figure*}

\Cref{fig:arg-quality-initial} displays the argument quality scores of all seven models on $50$ subsample of CMV sociopolitical posts (CMV$^{50}$). The metrics can be found in \Cref{tab:arg-quality-dims}. The higher the score, the better the argument quality. The black bars display the variance in quality across inference settings (nucleus sample decoding with temp=.9 and top\_p=.6, and greedy decoding). As mentioned earlier, the maximum number of tokens is $600$.

\paragraph{Results} \llamathree models generate substantially lower quality arguments compared to the rest. Out of the remaining top five, the overall winner is the \mistral. The scores from \llamatwo models are not significantly different (t-test with $\alpha$$<$$0.01$). Though significantly different on several dimensions (quality, clarity, cogency, reasonableness, and overall), \gptfour does not generate significantly dissimilar arguments in \emph{effectiveness} and \emph{impact}, which quantify the persuasiveness and the degree of likability of the arguments. Moreover, this model is approx. $15$ times more expensive and currently suffers from API-request timeouts due to high demand, making it approx. $9$ times slower than \gptthree.

\subsubsection{LLM Data}
No prompt engineering or optimization were employed in this study. For \llamatwoseven and \mistral, we use the HuggingFace text generation pipeline at \url{https://huggingface.co/docs/text-generation-inference}, and we access \gptthree via OpenAI text completion API at \url{https://platform.openai.com/docs/guides/gpt} (accessed between Dec. 8 - 19. 2024). All LLM comments were generated following the nucleus sample decoding introduced in \Cref{app:llm-selection}.

\begin{table*}[th!]
    \centering
    \scriptsize
    \resizebox{1\textwidth}{!}{
    \begin{tabular}{@{}l|l|l@{}}
        \toprule
        \textbf{Dimension} & \textbf{Description} & \textbf{Score}\\
        \midrule
        \textbf{Quality} & General argument quality score & ($0$-$1$)\\
        \textbf{Clarity} & Is it hard or easy to interpret the argument? & ($0$-$1$)\\
        \textbf{Impact} & User likes / recommendations multi-class (3 classes) & ($0$-$1$)\\
        \textbf{Overall} & General argument quality score & ($1$-$5$)\\
        \textbf{Cogency} & Acceptable and sufficient premises to draw a conclusion score & ($1$-$5$)\\
        \textbf{Reasonableness} & Contribution to the resolution of issues, the argument is accepted by universal audience score & ($1$-$5$)\\
        \textbf{Effectiveness} & Persuasion, rhetorical, emotional appeal score & ($1$-$5$)\\
        \midrule
    \end{tabular}
    }
    \caption{Argument quality dimensions and their respective score ranges in models from \citet{falk-lapesa-2023-bridging}.}
    \label{tab:arg-quality-dims}
\end{table*}

\subsubsection{Extracting Social Dimensions}
\label{app:social-dimensions}
\citet{choi2020TenDims} previously developed classifiers to estimate the likelihood that a message conveys a dimension $d$. Specifically, they trained a binary classifier (an LSTM) per dimension ($C_d$) on sentences to predict $p(d)$, which altogether can be seen as a multi-label classifier, as any sentence might convey several dimensions. For the CMV comments (analogously for LLM-generated text), we feed the input into the model a sentence at a time and obtain scores for each sentence in the text. We take the maximum score as the final output, specifically $s_d(m) = \text{max}_{S\in m}s_d(S)$, which then means that a message is considered to express a dimension as likely as the most likely sentence in the message.

\paragraph{Binarization and weight discounting} For consistency with prior work from \citet{monti2022language}, we do not majorly alter the notations in the following equations.

A message is considered to convey a dimension $d$ if $s_d(m)$ is above a threshold of $\theta_d$:
\begin{equation}
    d(m) =
    \begin{cases}
        1, \text{ if } s_d(m)\ge\theta_d\\
        0, \text{ otherwise}
    \end{cases}
\end{equation}

\noindent Finally, the weight discounted dimension score ($d(m)$) is: $\frac{1}{1+z\text{len}_d(m)}$ if $s_d(m) \ge \theta_d \land z\text{len}_d(m) \ge 0$, $2 - \frac{1}{1-z\text{len}_d(m)}$ if $s_d(m) \ge \theta_d \land z\text{len}_d(m) < 0$, and $0$ if $s_d(m) < \theta_d$.

\subsubsection{Measuring Social Intent and Opinion Change}
\label{app:odds-ratios}
The length-discounted prior probability of a message (post, human-written, or LLM-generated comment) being labeled with dimensions d is
\begin{equation}
    p(d) = \frac{\sum_{m\in M}d(m)}{2|M|},
\label{eq:prior-probability}
\end{equation}
where $M$ is the set of messages, and the factor of 2 is used to limit $p(d)$ between $0$ and $1$, as the values of $d(m)$ range from $0$ to $2$.

The conditional probability that a comment contains $d$, given that it received a $\Delta$ is
\begin{equation}
    p(d|\Delta) = \frac{\sum_{m\in C_\Delta}d(m)}{2|C_\Delta|},
\label{eq:delta-probability}
\end{equation}
where $C_\Delta$ is the set of $\Delta$ comment (analogously for $\overline{\Delta}$ and $LLM$). The odds ratio between $d$ in $\Delta$ and $\overline{\Delta}$ is defined in \Cref{eq:odds-ratios}:

\begin{equation*}
    \text{OR}(p(d|\Delta), p(d|\overline{\Delta})) = \frac{\text{odds}(p(d|\Delta))}{\text{odds}(p(d|\overline{\Delta}))},
\label{eq:odds-ratios-delta-nodelta}
\end{equation*}

\noindent where $\text{odds} = \frac{p}{1 - p}$ and $p(d|\Delta)$ is the conditional probability that a comment contains $d$, given that it received $\Delta$ (similarly for $\overline{\Delta}$). 
To compare human- and LLM-written comments, we use analogous $\text{OR}(p(d|\text{LLM}), p(d|\Delta))$ as:

\begin{equation}
    OR(p(d|LLM), p(d|\Delta))
    = \frac{\text{odds}(p(d|LLM))}{\text{odds}(p(d|\Delta))}.
\end{equation}

The conditional probability of a comment containing dimension $d_i$, given that its corresponding post contains dimension $d_j$ in the set of $\Delta$ (and equivalently for $\overline{\Delta}$ and $LLM$) comments is
\begin{equation}
    p_\Delta(d_i|d_j) = \frac{\sum_{c\in C_\Delta(P_{d_j})}d_i(c)}{2|C_\Delta(P_{d_j})|}.
\label{eq:cond-probability}
\end{equation}

The odds ratio between the conditional probabilities between $\Delta$ and all CMV comments ($\Delta$ and $\overline{\Delta}$) is defined in \Cref{eq:odds-ratios-cond}:

\begin{equation*}
    \text{OR}(p_\Delta(d_i|d_j), p(d))\\
    = \frac{\text{odds}(p_\Delta(d_i|d_j))}{\text{odds}(p(d))},
\label{eq:odds-ratios-cond-app}
\end{equation*}
with a $95\%$ confidence interval defined as 

\begin{equation}
    ci = 1.96 \sqrt{\frac{1}{|C_{d,\Delta}|}+\frac{1}{|C_{d,\overline{\Delta}}|}}.
\label{eq:confidence-interval}
\end{equation}

Analogously, the odds ratio between the conditional probabilities LLM-generated and all CMV comments ($\Delta$ and $\overline{\Delta}$) is defined as

\begin{equation}
    \text{OR}(p_{LLM}(d_i|d_j), p(d))\\
    = \frac{\text{odds}(p_{LLM}(d_i|d_j))}{\text{odds}(p(d))},
\label{eq:odds-ratios-cond-llm}
\end{equation}
where $p(d)$ is the prior probability of dimensions in CMV comments, and the confidence interval used in \Cref{eq:odds-ratios-cond-llm} includes this additional term $\frac{1}{|C_{d,LLM}|}$ to control for the model-generated texts.

\section{Crowdsourcing}
\label{sec:appendix_b}
This section provides details about the crowdsourcing.
\paragraph{Annotation}
Annotators were shown a single annotation instance in each page -- a post with its title, a human-written $\Delta$ comment, and a \gptthree-generated comment. Below each post, we asked \emph{``Do you agree with this message?''} with options ``Yes'' and ``No''. The preselected ``N/A'' was not a valid answer, i.e., the annotators could not move to the next page unless they answered the questions. Comments were displayed in collapsible sections, with the same question and answer options below each comment.
At the bottom of each page, we asked the annotators \emph{``Which of the messages is more likely to change the opinion holder's view?''} Annotators had to select ``Message 1'' or ``Message 2''. Importantly, the comments were displayed in random orders, meaning that we randomly rotated which comment would be displayed first, to avoid biasing the annotators. 

As required by the survey platform, the annotators were given two simple attention checks. The crowdworkers who failed the attention checks have been excluded from the study.

\paragraph{Compensation} The crowdworkers were compensated at a rate of £12/h, which is well above the federal minimum wage for covered nonexempt employees of £5,76/h (or \$7.25 USD).

\paragraph{Annotator demographics}
The worker pool consisted of U.S. residents who were eighteen years old or above and fluent in English (with a fluency level of either fluent or native speaker). The annotators were balanced by gender (20 female and 20 male) and consisted of both U.S. nationals and immigrants with diverse ethnic -- 6 Asian, 11 Black, 18 White, 3 Mixed, 2 non-disclosed -- educational, and employment backgrounds. Countries of birth included predominantly USA, with one or two participants from countries such as Libya, Nigeria, India, Germany, Mexico, Ghana, Algeria, Philippines, and China. They were warned about potentially triggering conversation topics such as violence, discrimination, or suicide before asking for consent to continue the annotation with an option to drop out of the study and withdraw consent at any point.

\begin{table}[t]
    \centering
    \scriptsize
    \resizebox{0.5\textwidth}{!}{
    \begin{tabular}{@{}l|c|c|c@{}}
    \multicolumn{2}{l}{} & \multicolumn{2}{c}{\textbf{Preferred}} \\
        \cmidrule(lr){3-4}
    \multicolumn{1}{l|}{\textbf{Agreed with}} & \multicolumn{1}{c|}{$\#$ Cases} & \multicolumn{1}{c}{$\Delta$ Comment} & \multicolumn{1}{c}{GPT Comment}\\
        \midrule
        \textbf{Both / Neither} & 58 & 9 & 49 \\
        \textbf{$\Delta$ Comment} & 9 & 6 & 3\\
        \textbf{GPT Comment} & 33 & 2 & 31\\
    \end{tabular}
    }
    \caption{Annotator preferences based on their agreement with the comment from $\Delta$ or \gptthree. Note that the annotator \emph{can} agree with both arguments in a single instance as they do not choose which one they agree with but indicate \emph{whether they agree with} each text (post, human-written $\Delta$ comment, and \gptthree-generated comment).}
    \label{tab:stance-agreement}
\end{table}

\section{Further Results}
\label{sec:appendix_c}
This section provides further results.

\begin{figure*}[t]
    \begin{subfigure}[t]{0.32\textwidth}
        \centering
        \includegraphics[height=2in]{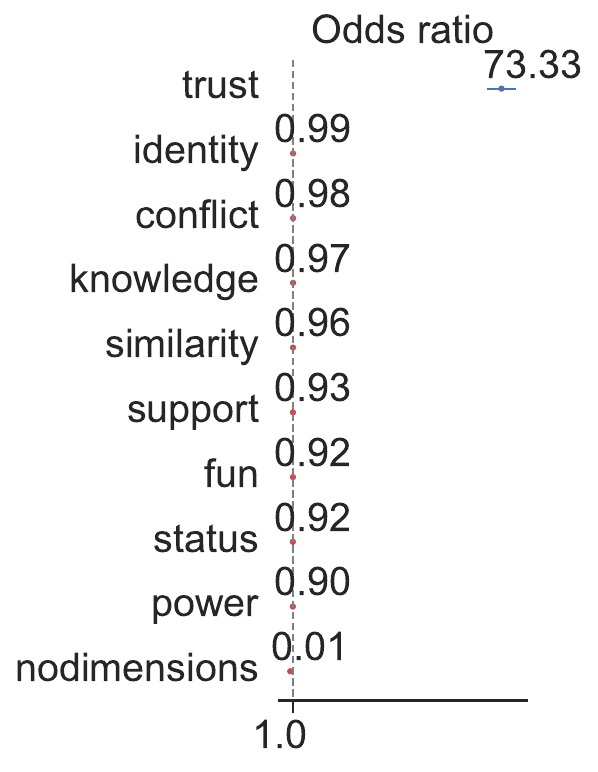}
        \caption{\label{fig:odds-llama-human} \llamatwoseven vs. CMV}
    \end{subfigure}%
    \begin{subfigure}[t]{0.32\textwidth}
        \centering
        \includegraphics[height=2in]{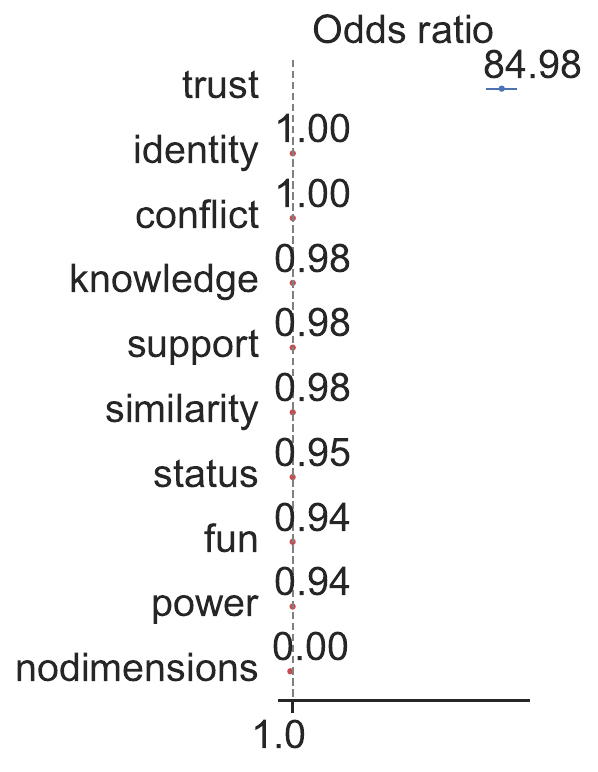}
        \caption{\label{fig:odds-mistral-human} \textsc{Mistral}-7B vs. CMV}
    \end{subfigure}%
    \begin{subfigure}[t]{0.32\textwidth}
        \centering
        \includegraphics[height=2in]{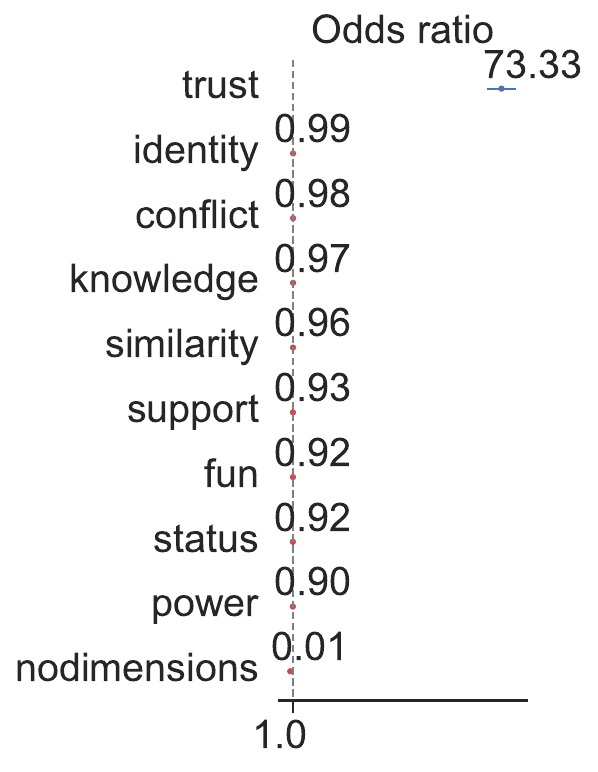}
        \caption{\label{fig:odds-gpt-human} \gptthree vs. CMV}
    \end{subfigure}
    \caption{Odds ratios between probabilities of social dimensions present in texts measured as in \Cref{eq:odds-ratios} -- comparison of LLM-generated and CMV comments ($\Delta$ and $\overline{\Delta}$). Error bars represent the 95\% confidence intervals (see \Cref{eq:confidence-interval} and details in \Cref{app:odds-ratios}). The difference between the LLM-generated and all human-written comments is minimal except for the \emph{trust} and \emph{nodimensions}.}
    \label{fig:odds-ratios-all-human}
\end{figure*}

\paragraph{LLMs vs. CMV Comments}
\Cref{fig:odds-ratios-all-human} presents the odds ratios comparing the likelihood that LLMs, relative to CMV commenters ($\Delta$ and $\overline{\Delta}$), generate texts expressing a given social dimension $d$. Notably, the models are substantially more likely than human commenters to express \emph{trust} toward their interlocutors and are rarely observed to not express any illocutionary intent.

While differences are less pronounced across the remaining dimensions, notable disparities can be observed in \emph{support}, \emph{fun}, \emph{status}, and \emph{power}. These dimensions exhibit the largest gaps among the remaining dimensions, with some variation across models. For instance, \emph{support}, \emph{fun}, and \emph{status} are approximately 8\% less likely to be generated by the chat-optimized models (\llamatwoseven and \gptthree), whereas this trend is not as evident for the instruction-tuned model (\mistral). This contrast suggests that alignment training may play a role in shaping the communicative behaviors exhibited by different LLMs.

\end{document}